\documentclass[conference]{IEEEtran}
\IEEEoverridecommandlockouts
\usepackage{cite}
\usepackage{amsmath,amssymb,amsfonts}
\usepackage{algpseudocode}
\usepackage[ruled]{algorithm2e}
\usepackage{graphicx}
\usepackage{textcomp}
\usepackage{xcolor}
\usepackage{bm}
\usepackage{amsthm}
\usepackage{xstring}
\usepackage{xfrac}
\usepackage[hidelinks]{hyperref}

\def\BibTeX{{\rm B\kern-.05em{\sc i\kern-.025em b}\kern-.08em
    T\kern-.1667em\lower.7ex\hbox{E}\kern-.125emX}}

\newcommand{\x}[0]{\mathbf{x}}
\newcommand{\y}[0]{\mathbf{y}}

\newcommand{\cc}[0]{\mathbf{c}}

\begin{document}

\theoremstyle{definition}
\newtheorem{defn}{Definition}

\title{Visualizing Riemannian data with Rie-SNE}

\author{\IEEEauthorblockN{Andri Bergsson}
\IEEEauthorblockA{\textit{PayAnalytics (\href{https://www.payanalytics.com/}{payanalytics.com})}\\
Reykjavík, Iceland \\
and.bergsson@gmail.com}
\and
\IEEEauthorblockN{Søren Hauberg}
\IEEEauthorblockA{\textit{Technical University of Denmark} \\
\textit{Kgs. Lyngby, Denmark}\\
sohau@dtu.dk}
}

\maketitle

\begin{abstract}
  Faithful visualizations of data residing on manifolds must take the underlying geometry into account when producing a flat planar view of the data. In this paper, we extend the classic \emph{stochastic neighbor embedding (SNE)} algorithm to data on general Riemannian manifolds. We replace standard Gaussian assumptions with Riemannian diffusion counterparts and propose an efficient approximation that only requires access to calculations of Riemannian distances and volumes. We demonstrate that the approach also allows for mapping data from one manifold to another, e.g.\@ from a high-dimensional sphere to a low-dimensional one.
\end{abstract}


\section{Introduction}\label{sec:intro}
  Visualizations are crucial to investigators trying to make sense of high-dimensional data. The most common output of a visualization is a two-dimensional plot (e.g.\@ on a piece of paper or a computer screen), so we often call on a form of dimensionality reduction when working with high-dimensional data. The vast majority of dimensionality reduction techniques assume that data resides on a Euclidean domain (Sec.~\ref{sec:euc_viz}), which presents a problem when data is not quite that simple. Data residing on Riemannian manifolds, such as the sphere, appear in many domains where either known constraints or other modeling assumptions impose a Riemannian structure (Sec.~\ref{sec:riem_data}). In such settings, how should one visualize data?
  
  There are many concerns and questions when visualizing Riemannian data. The first is generic: all dimensionality reduction tools amplify parts of the signal, while reducing the remainder. This is an inherent limitation, which should always be in mind when interpreting data visualizations. Since some loss of information is inevitable, should we then loosen our grip on the data or its underlying Riemannian structure when such is present? Gauss's \emph{Theorema Egregium} \cite{gauss2005general} informs us that if the final plot is to be presented on a \emph{flat} screen or piece of paper, then a distortion of the Riemannian structure is inevitable.
  
  In practice, even if one accepts the limitations of a visualization, actual algorithms for visualizing Riemannian data are missing. In this paper, we develop an extension of the \emph{Stochastic Neighbor Embedding} \cite{sne} method to Riemannian data and thereby provide one such tool. We call this \emph{Riemannian Stochastic Neighbor Embedding}, or \emph{Rie-SNE} for short. Our approach is quite general as it allows for embedding data observed on one Riemannian manifold to be embedded on another. This allows for mapping data from a Riemannian space to a two-dimensional Euclidean plane (for plotting), but also mapping to a two-dimensional sphere, or similar, when the Euclidean topology is inappropriate. Rie-SNE does not claim to solve the above-mentioned limitations of visualization, but it does provide a working tool, which we demonstrate to have practical merit. 
  
  \begin{figure}[t]
      \centering
      \includegraphics[width=0.9\columnwidth]{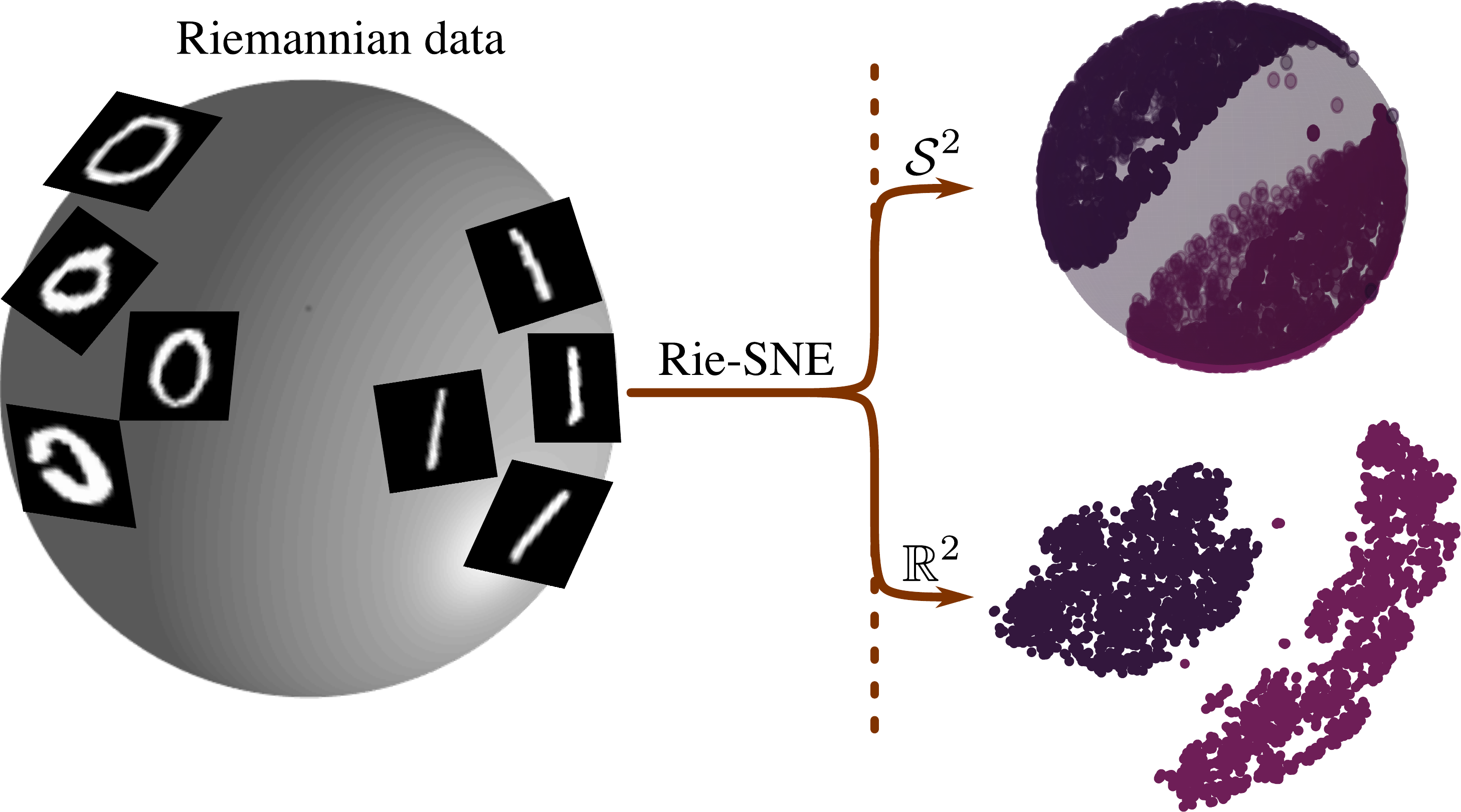}
      \caption{Rie-SNE visualizes high-dimensional Riemannian data by mapping to a low-dimensional Riemannian (or Euclidean) manifold, which can then be shown. The figure shows high-dimensional spherical data mapped to either a low-dimensional sphere or a low-dimensional plane. The method also supports other manifolds, both as input and output.}
      \label{fig:teaser}
  \end{figure}

\section{Background and related work}\label{sec:background}
  Before describing Rie-SNE, we provide the relevant context on visualization and geometry.

  \subsection{Euclidean visualization}\label{sec:euc_viz}
    General data visualization is a vast topic, and a complete review is beyond our scope \cite{vanderplas2016python}. We here focus on the setting where data is represented as vectors (points) in a Euclidean space of high dimension. When data is two- or three-dimensional a scatter plot can directly reveal its structure, and we focus on the more difficult setting where data dimensions vastly exceed the easily plottable. Here one may explore the data through multiple projections, such as pairwise scatter plots, which is usually manageable for data of up to around 10 dimensions. Eventually the approach tends to become unwieldy and the greater picture is lost. Alternatives include continuously interpolating between dimensions to provide an (interactive) animation \cite{asimov1985grand}.
  
    The approach we here explore is to find a non-linear mapping from the high-dimensional observation space into a two- or three-dimensional space, which is suitable for plotting. Many variants of this approach exists, and we only touch upon a few. \emph{Principal component analysis (PCA)} \cite{jolliffe2002principal} is perhaps the most commonly used approach to dimensionality reduction. This seeks a low-dimensional representation of data that preserves as much variance as possible. The restriction to spanning a linear subspace of the observation space, however, often implies that the low-dimensional view reveal little structure. The \emph{Gaussian process latent variable model (GP-LVM)} \cite{gplvm} provides a nonlinear probabilistic extension of PCA that places a Gaussian process prior on the unknown mapping that reduce dimensionality, and marginalize this accordingly. The approach carries intrinsic elegance, but its optimization can be brittle \cite{feldager:icann:2021}. Classic `manifold learning' techniques avoid optimization issues by phrasing optimization tasks, where an optimum is available through spectral decompositions \cite{isomap, lle, laplacian}. These rely on constructions of neighborhood graphs, which, however, can be brittle, so in practice the resulting visualizations are sensitive to parameter choices. The \emph{stochastic neighbor embedding (SNE)} \cite{sne} replaces the `hard' graph construction with a softer construction. A modern variant of this method \cite{vanDerMaaten2008} is one of the currently most popular algorithms, and is the one we here extend. We cover this in Secs.~\ref{sec:sne} and \ref{sec:tsne}.
    
  \subsection{Riemannian data}\label{sec:riem_data}
    Data is often equipped with additional knowledge such as constraints or given smooth structures. In many scenarios this additional knowledge gives the data a Riemannian structure. For example, knowing that all observations have unit norm, places them on the unit sphere, which has a well-studied Riemannian structure. The a priori available knowledge giving rise to a Riemannian data interpretation differs between domains, and we here only name a few. The most prominent example is that of \emph{directional statistics} \cite{mardia2000directional, kurz2013recursive, hauberg:SN:2018} where data resides on the unit sphere. Other examples include \emph{shape data} \cite{Kendall:BLMS:1984, Freifeld:ECCV:2012, Srivastava:PAMI:2005, Younes:ICV:2012, kurtek:pami:2012}, \emph{DTI images} \cite{lenglet:eccv:2004, Pennec:IJCV:2006,wang2014tracking}, \emph{image features} \cite{Tuzel:ECCV:2006, Porikli:CVPR:2006, freifeld:cvpr:2014}, \emph{motion models}~\cite{turaga,Cetingul:cvpr:2009}, \emph{human poses} \cite{said:eusipco:2007, Hauberg:IMAVIS:2011, spatial_priors:hauberg_et_al10, hauberg:mukf}, \emph{robotics} \cite{gilitschenski2015unscented, jaquier2022geometry}, \emph{social networks} \cite{mathieu2019continuous} and more.
  
    Even if Riemannian data is becoming increasingly common, tools for visualization have not followed. The most common approach for visualizing Riemannian data is to locate a point on the manifold, e.g.\@ the intrinsic mean \cite{pennec}, and map the data to the tangent space of this point. That gives a Euclidean view of the data, which can then be visualized with one of the many methods for this domain (Sec.~\ref{sec:euc_viz}). In particular, applying PCA tangentially is the gold standard \cite{fletcher2004principal}. Unless data is concentrated around the point of tangency this approach is bound to give a highly distorted view of the data, and in practice most knowledge reflected in the geometry will be lost by the linearization \cite{sommer:eccv10}. Several extensions of PCA to Riemannian manifolds exist, e.g.\@ \cite{huckemann,  panaretos:jasa:2014, jung:biometrika:2012}. These focus on generalizing the classic linear models, and less work has been done on extending nonlinear methods. Two notable extensions are a `wrapped' extension of the GP-LVM \cite{mallasto:2018}, and an extension of the classic principal curves model \cite{hauberg:tpami:princurve}.

  \subsection{Measuring on Riemannian manifolds}\label{sec:geom}
    In order to engage with data residing on a manifold, we need a collection operators that can be applied to data. Here we only cover the most basic as that is all required by Rie-SNE. A detailed exposition can be found elsewhere \cite{pennec, doCarmoRiemannian}.
    
    A Riemannian manifold is a space which is locally Euclidean. This imply that in a neighborhood around a point $\bm{\mu}$ on the manifold, we can get a Euclidean view of the manifold in the form of a tangent space, which is equipped with an inner product,
    \begin{align}
      \langle \x_i, \x_j \rangle_{\bm{\mu}} = \x_i^\top G_{\bm{\mu}} \x_j,
    \end{align}
    where $G_{\bm{\mu}}$ is a symmetric positive definite matrix that reflects the inner product at $\bm{\mu}$. This inner product is allowed to change smoothly between tangent spaces, in order to compensates for the approximation error induced by linear view of the manifold. This distortion can be characterized by the change-in-volume between the manifold and its tangent, which follows $\sqrt{\det G_{\bm{\mu}}}$ \cite{pennec}.
    
    The inner product allow us to define local distances, which can be integrated to provide a notion of \emph{curve length}. That is, given a curve $\cc$ on a manifold, we may compute its length as
    \begin{align}
      \mathrm{Length}[\cc] &= \int \sqrt{\langle \dot{\cc}_t, \dot{\cc}_t \rangle}_{\cc_t} \mathrm{d}t,
      \label{curvelength}
    \end{align}
    where we assume the curve to be parametrized by $t \in [0, 1]$, and use $\cc_t$ and $\dot{\cc}_t$ to denote the position and velocity of the curve, respectively.
    From the notion of a curve length, it is trivial to define the distance between two points as the length of the shortest curve,
    \begin{align}
      \mathrm{dist} &= \min_{\cc} \mathrm{Length}[\cc].
      \label{geodesic}
    \end{align}
    The shortest curve commonly goes under the name \emph{geodesic}.
    The distance function can be differentiated using the relation
    \begin{align}
      \partial_{\x_i} \mathrm{dist}^2(\x_i, \x_j) &= 2 \mathrm{Log}_{\x_i}(\x_j),
    \end{align}
    where $\mathrm{Log}_{\x_i}(\x_j)$ is the Riemannian \emph{logarithm map} \cite{pennec}. If $\cc$ is a constant-speed geodesic connecting $\x_i$ and $\x_j$, then the logarithm map is merely the initial velocity $\dot{\cc}_0$ of said curve.
    
  \subsection{Stochastic neighbor embedding}\label{sec:sne}
    \emph{Stochastic neighbor embedding (SNE)} \cite{sne} is a dimensionality reduction tool, which aims to preserve similarity between neighboring points when mapped to a low-dimensional representation. Assume for now, that we have access to function $s_{\text{high}}$
    and $s_{\text{low}}$, which measure the similarity between observation pairs in the high-dimensional observation space and the low-dimensional representation space, respectively.
    Now define the conditional probability, $p_{j|i}$ that $\x_i$ would pick $\x_j$ as its neighbor \cite{vanDerMaaten2008} 
    \begin{align}
      p_{j|i} = \frac{s_{\text{high}}(\x_j | \x_i)}{\sum_{k \ne i} s_{\text{high}}(\x_k | \x_i)}.
      \label{hdimp2}
    \end{align}
    Common convention is to define $p{i|i} = 0$. Further note that $\sum_j p_{j|i} = 1$. We can renormalize this to form a distribution over all observations as
    \begin{gather}
      p_{ij} = \frac{p_{j|i} + p_{i|j}}{2n},
      \label{hdimp1}
    \end{gather}
    where $n$ is the number of observations.
    
    To learn a low-dimensional representation, the key idea is to repeat the above over the low-dimensional space to form
    \begin{align}
      q_{j|i} &= \frac{s_{\text{low}}(\y_j | \y_i)}{\sum_{k \ne i} s_{\text{low}}(\y_k | \y_i)}, \\
      q_{ij}  &= \frac{q_{j|i} + q_{i|j}}{2n}.
    \end{align}
    We can now compare the similarity of our data and the representation by computing the Kullback-Leibler divergence between $p_{ij}$ and $q_{ij}$,
    \begin{align}
      C = \mathrm{KL}\left( P || Q \right) = \sum_{i=1}^n \sum_{j=1}^n p_{ij} \log \frac{p_{ij}}{q_{ij}}.
    \end{align}
    This can then be minimized using gradient descent with respect to the low-dimensional representation $\{ \y_i \}_{i=1}^n$.

    In its classic form, SNE picks the measures of similarity as Gaussian functions
    \begin{align}
    \begin{split}
      s_{\text{high}}(\x_j | \x_i)
        &= \left(2\pi\sigma_i^2\right)^{-\sfrac{D}{2}} \exp\left( -\frac{\| \x_j - \x_i \|^2}{2\sigma_i^2} \right), \\
      s_{\text{low}}(\y_j | \y_i)
        &= \left(2\pi\right)^{-\sfrac{d}{2}} \exp\left( -\frac{\| \y_j - \y_i \|^2}{2} \right).
    \end{split}\label{eq:sne_s}
    \end{align}
    With this choice, the normalization constants of Eq.~\ref{eq:sne_s} cancel out when computing $p_{j|i}$ and $q_{j|i}$.
    Note that this approach gives a per-observation variance $\sigma_i^2$, such that different points effectively can have different sizes of neighborhoods. To determine the $\sigma_i^2$ parameters, the user specifies a \emph{perplexity} parameter, which can be thought of as a measure of the effective number of neighbors \cite{vanDerMaaten2008}. This is defined as
    \begin{align}
      \text{perplexity} = 2^{H\left(P_i\right)},
      \label{eq:perplexity}
    \end{align}
    where $H\left(P_i\right)$ is the Shannon entropy \cite{journals/bstj/Shannon48} of $P_i$ in bits:
    \begin{align}
      H\left(P_i\right) = - \sum_{j} p_{j|i} \log_2 p_{j|i}.
    \end{align}
    For a specific user-provided value of the perplexity parameter, we can perform a binary search over $\sigma_i^2$ such that Eq.~\ref{eq:perplexity} holds. In practice, the user experiments with different choices of perplexities to see which reveals a pattern.

  \subsection{The t-distributed stochastic neighbor embedding}\label{sec:tsne}
    The most popular variant of SNE is the \emph{t-distributed SNE} \cite{vanDerMaaten2008}. This is motivated by the so-called `crowding problem' often observed in SNE, where the low-dimensional representations significantly overlap without revealing much underlying structure. The idea is to use a similarity in representation space with more heavy tails than the Gaussian. Specifically, $s_{\text{low}}$ is chosen as a $t$-distribution with one degree of freedom centered around one representation, i.e.
    \begin{align}
      s_{\text{low}}(\y_j | \y_i)
        &= \pi^{-1} \left( 1 + \|\y_j - \y_i\|^2 \right)^{-1}.
    \end{align}

  As is evident, t-SNE needs to compute all pairwise distances between data points and therefore has quadratic complexity. Using approximation techniques, such as vantage-point trees or the \textit{Barnes-Hut approximation}, the running time can be lowered down to having $\mathcal{O}(n \log n)$ complexity \cite{tsneOptimization}.

  \begin{figure*} 
      \centering
      \includegraphics[width=\textwidth]{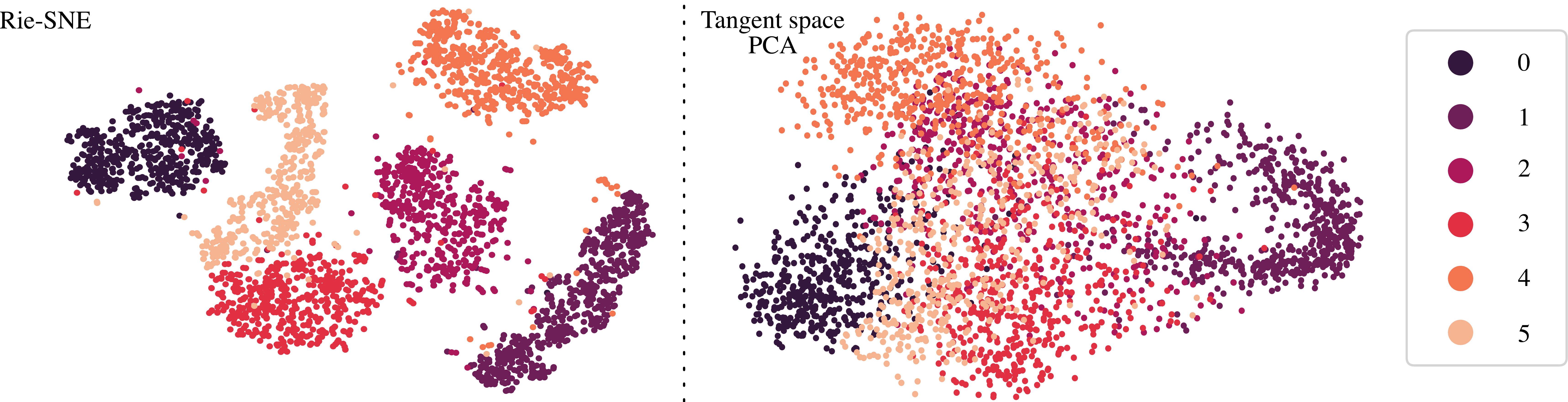}
      \caption{Two-dimensional Euclidean embeddings of spherical MNIST. The left panel show the embedding obtained by Rie-SNE while the right panel show the gold standard of tangent space PCA. Note how Rie-SNE discovers structure, which is lost on tangent space PCA.}
      \label{fig:mnist6_plane}
  \end{figure*}

\section{Method}
  The inner workings of Rie-SNE, or \emph{Riemannian Stochastic Neighbor Embedding}, are now examined.

  \subsection{Brownian motion on a Riemannian manifold}
    The key building block for generalizing SNE to Riemannian manifolds is a suitable generalization of the Gaussian distribution. Here we consider a density derived from a Brownian motion on a Riemannian manifold for high-dimensional probability computations.

    Given a Brownian motion in Euclidean space, the probability that the random walk will end in point $\x$ can be computed by using the Gaussian density. If the increments of the Brownian motion are sufficiently small, each increment can be projected onto the tangent space of corresponding points on a Riemannian manifold without error. Then, a Brownian motion starting at point $\bm{\lambda}$ running for some time $t$ can be projected onto a $D$-dimensional Riemannian manifold and there it will give rise to a random variable, a random variable that can be interpreted as the probability that a Brownian motion starting at $\bm{\lambda}$ will end in point $\x$ on the manifold. Since now the Brownian motion has been projected onto a Riemannian manifold the density will be different, and it can be approximated with a paramatrix expansion \cite{hsu2002stochastic,kalatzis2020variational}:
    \begin{align}
      \mathcal{BM}(\x|\bm{\lambda}, t) &\approx
        \left( 2 \pi t \right)^{-\frac{D}{2}}H_{0} \exp \left( - \frac{\mathrm{dist}^2 (\x, \bm{\lambda})}{2t}\right)
      \label{eq:riemannianbrownianmotion}
    \end{align}
    where
    \begin{itemize}
      \item [-] $t \in \mathbb{R}_+$ is the duration of the Brownian motion and corresponds to variance in Euclidean space.
      \item [-] $\bm{\lambda}$ is the starting point of the Brownian motion.
      \item [-] $H_0$ is the ratio of Riemannian volume measures evaluated at points $\x$ and $\bm{\lambda}$ respectively, i.e.:
        \begin{align}
          H_0 = \left( \frac{\det G_{\x}}{\det G_{\bm{\lambda}}}\right)^{\frac{1}{2}}
        \end{align}
        with again $G_{\mathbf{p}}$ being the metric evaluated at $\mathbf{p}$.
    \end{itemize}
    Superficially, Eq.~\ref{eq:riemannianbrownianmotion} looks like the density of the normal distribution, with the Euclidean distance being replaced by its Riemannian counterpart. However, here it is worth noting that the normalization factor $H_0$ is different from the usual Euclidean distribution.

  \subsection{Rie-SNE}
    \emph{Rie-SNE} works in a similar manner as SNE and t-SNE, i.e.\@ it will also produce two probability distributions $P$ and $Q$ from the data and aim to make them as similar as possible and in the process capture some underlying structure in the produced low dimensional embedding. However, computing the high-dimensional probability distribution $P$ comes with an added cost. To preserve the Riemannian nature of the data, a different density is used when computing high-dimensional probabilities belonging to $P$, namely the approximate density induced by the heat kernel of a Brownian motion on a Riemannian manifold given in Eq.~\ref{eq:riemannianbrownianmotion}, i.e.\@ we pick
    \begin{align}
      s_{\text{high}}(\x_j|\x_i) &= \mathcal{BM}{\x_j | \x_i, t_i}.
    \end{align}
    The added computational cost is that the evaluation of $\mathcal{BM}(\cdot|\cdot)$ is more demanding than the conventional Gaussian similarity \eqref{eq:sne_s}. Specifically, the normalization $H_0$ and the geodesic distance may be demanding, depending on the manifold on which the data resides.

    With the Browninan motion model we get
    \begin{align}
      p_{j|i} \approx \frac{
        H_{0}[i,j] \cdot  \exp \left( - \frac{\mathrm{dist}^2[i,j]}{2t_i}\right)
      }{\sum\limits_{k \ne i} H_{0}[i,k] \cdot \exp \left( - \frac{\mathrm{dist}^2[i,k]}{2t_i}\right)},
    \end{align}
    where we use the notations $\mathrm{dist}^2[i,j] = \mathrm{dist}^2(\x_i,\x_j)$ and $H_{0}[i,j] = \sqrt{ \sfrac{\det G_{\x_i}}{\det G_{\x_j}}}$ to emphasize that these quantities can be pre-computed. As with SNE, we can optimize $t_i$ to match a pre-specified perplexity using a binary search.
    
  \begin{figure*} 
      \includegraphics[width=0.95\textwidth]{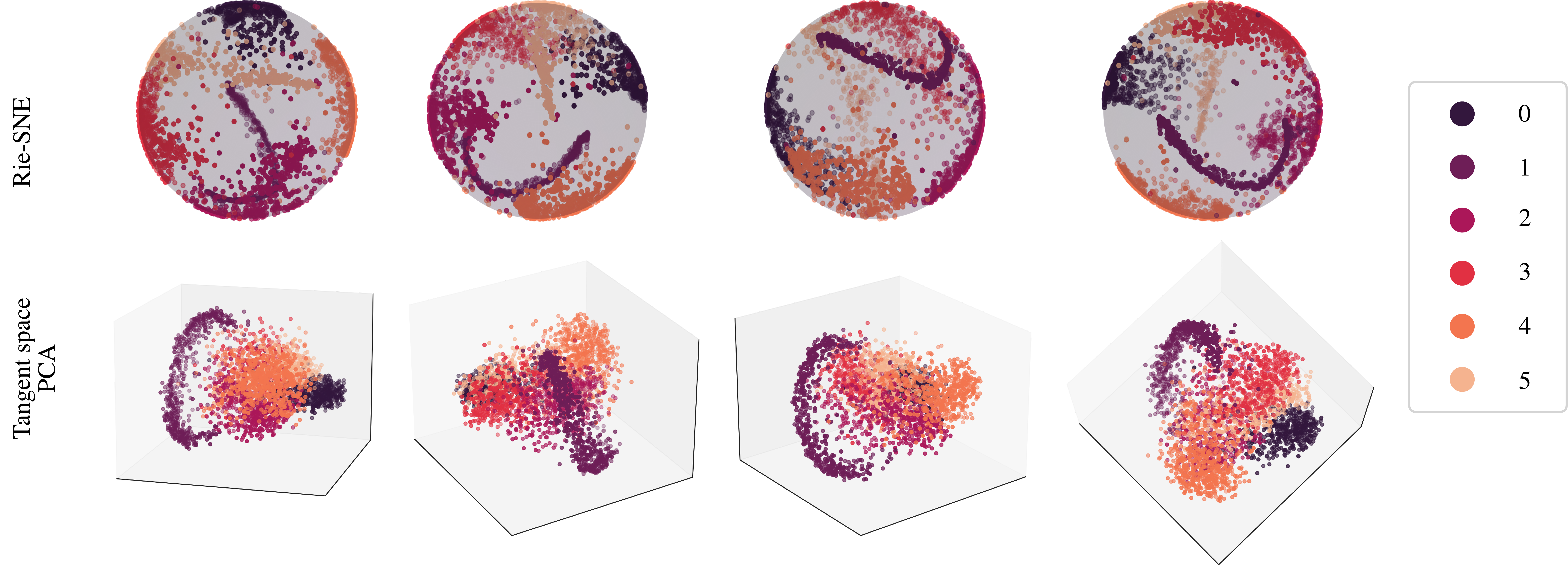}
      \vspace{-3mm}
      \caption{Four different views of embeddings of spherical MNIST. The top row show spherical embeddings obtained using Rie-SNE, while the bottom row show three-dimensional embeddings using the gold standard tangent space PCA. Note that the clustering structure is significantly more evident in Rie-SNE.}
      \label{fig:mnist6}
  \end{figure*}

  \subsection{Choice of representation}
    As mentioned in Sec.~\ref{sec:intro}, Gauss's \emph{Theorema Egregium} \cite{gauss2005general} inform us that we cannot isometrically embed data from a curved space into a space of different curvature without introducing distortion. Specifically, if we embed data from a nonlinear manifold onto a \emph{flat} two-dimensional representation (for plotting) then the curvature mismatch between spaces induces a distortion. This is a fundamental limitation that any visualization of Riemannian data will face, but we may nonetheless try to limit its impact. One approach is to embed the data onto a manifold of similar curvature as that of the manifold on which the data resides. For example, if the data resides on a high-dimensional sphere, it is perhaps more prudent to embed onto a two-dimensional sphere for plotting, rather than a Euclidean space.
    
    With this in mind, we choose different distributions over the low-dimensional representation, depending on user preference. 
    \begin{itemize}
      \item \textbf{Euclidean.} If the user prefers a Euclidean low-dimensional representation, we opt to use a student-t as in regular t-SNE,
      \begin{align}
        s_{\text{low}}(\y_j | \y_i)
          &= \pi^{-1} \left( 1 + \|\y_j - \y_i\|^2 \right)^{-1}.
      \end{align}
      \item \textbf{Spherical.} If the data manifold has positive curvature it may be beneficial to embed on a sphere, in which case we opt to use a von Mises-Fisher distribution \cite{mardia2000directional},
      \begin{align}
        s_{\text{low}}(\y_j | \y_i)
          &= \left( \sqrt{2\pi} I_{\sfrac{d}{2}-1}(1) \right)^{-1} \exp\left( \y_j^{\top} \y_i \right),
      \end{align}
      where $I_{v}$ is a modified Bessel function of the first kind of order $v$. In practise the normalization constant cancels out and can be ignored.
      \item \textbf{Other.} The user may have other prior knowledge about the manifold on which the data resides, which may suggest embedding on some other low-dimensional manifold. In this case, we suggest to also use the Riemannian Brownian over the low-dimensional representation, i.e.,
      \begin{align}
        s_{\text{low}}(\y_j | \y_i)
          &= \mathcal{BM}(\y_j | \y_i, 1).
      \end{align}
    \end{itemize}
    
    Once we have defined both $s_{\text{high}}$ and $s_{\text{low}}$, we can estimate the representations using gradient descent just as regular SNE. Having performed $T$ iterations (with a sufficiently large value for $T$) of the gradient descent, the two probability distributions $P$ and $Q$ will have a minimal KL-divergence resulting in near-optimal positions of the points in the low-dimensional embedding. The key elements of the resulting computations are provided in Alg.~\ref{alg:main} on page~\pageref{alg:main}.
    
  \subsection{Implementation details}
    Our implementation of Rie-SNE relies on two approximation techniques that are traditionally also used when performing regular t-SNE. 
    
    First, we compute the high-dimensional probabilities in $P$ by using a sparse nearest neighbor-based approximation technique \cite{tsneOptimization}. This means we compute a sparse approximate $P$ distribution, where far-away points are given a probability of zero. This can be realized with a nearest neighbor search. Empirical results from van der Maaten \cite{tsneOptimization} suggest a value of $\tau = \lfloor 3 \cdot \text{perplexity} \rfloor$ nearest neighbors will give sufficiently good approximations of $P$, which we also use here. Finding the $\tau$ nearest neighbors of each point can be done by constructing a vantage-point tree \cite{vptrees} over the data and performing a nearest neighbors search on the resulting tree. Constructing the tree, performing the nearest neighbor search and computing the relevant values of $P$ has time complexity $\mathcal{O}(\tau n \log n)$.
    
    Second, we use the \emph{Barnes-Hut approximation} \cite{barnes86a}, whenever we opt to use a student's t-distribution over the low-dimensional representation. Minimizing the KL-divergence between the two probability distributions $P$ and $Q$ requires using gradient descent, and in each step of the gradient descent we need to compute all pairwise $q_{ij}$ of $Q$, which has quadratic complexity. The Barnes-Hut approximation of the gradient instead has complexity $\mathcal{O}(n \log n)$. In short, this approximation split the low-dimensional representation into quadrants (via tree structures named quadtrees/octtrees), such that points in far-away and small enough quadrants can be approximated as the same point appearing multiple times. For each low-dimensional point, a depth-first search with complexity $\mathcal{O}(\log n)$ is done to mark quadrants as approximate quadrants or not. A total of $n$ depth-first searches are carried out, yielding the $\mathcal{O}(n \log n)$ time complexity. 

\section{Results}\label{sec:results}
 The performance of Rie-SNE is shown by comparing it to the gold-standard of visualizing non-euclidean data. This amounts to first computing the intrinsic mean, mapping all data to the tangent space at this point, and performing PCA over the tangential data.
  
  \subsection{Spherical MNIST}
  We start with the classic MNIST dataset \cite{lecun-mnisthandwrittendigit-2010} consiting of $24 \times 24$ dimensional gray-scale images. We consider digits 0--5 to reduce clutter. To induce a non-Euclidean data geometry, we project the data onto the unit sphere of $\mathbb{R}^{24 \times 24}$ and denote the resulting data \emph{spherical MNIST}. We visualize the resulting data using both Rie-SNE and tangent space PCA. First, we embed the data onto the plane, $\mathbb{R}^2$, and show the resulting plots in Fig.~\ref{fig:mnist6_plane}. Here it can be seen that Rie-SNE captures well the underlying relationship between the data points (same digits are grouped together), while tangent space PCA produces a cluttered view which does not reveal the underlying structure. Since the data has a spherical geometry, it may be beneficial to embed onto a low-dimensional sphere to better preserve topology and curvature. Figure~\ref{fig:mnist6} show the Rie-SNE embedding onto $\mathcal{S}^2$, where the clustering is again evident. As a baseline, the figure also shows a tangent space PCA embedding on $\mathbb{R}^3$. Although some structure can be captured with tangent space PCA here, Rie-SNE still gives better separation of the digit classes.
  
  \subsection{Crypto-tensors}
\begin{figure} 
      \centering
      \includegraphics[width=\columnwidth]{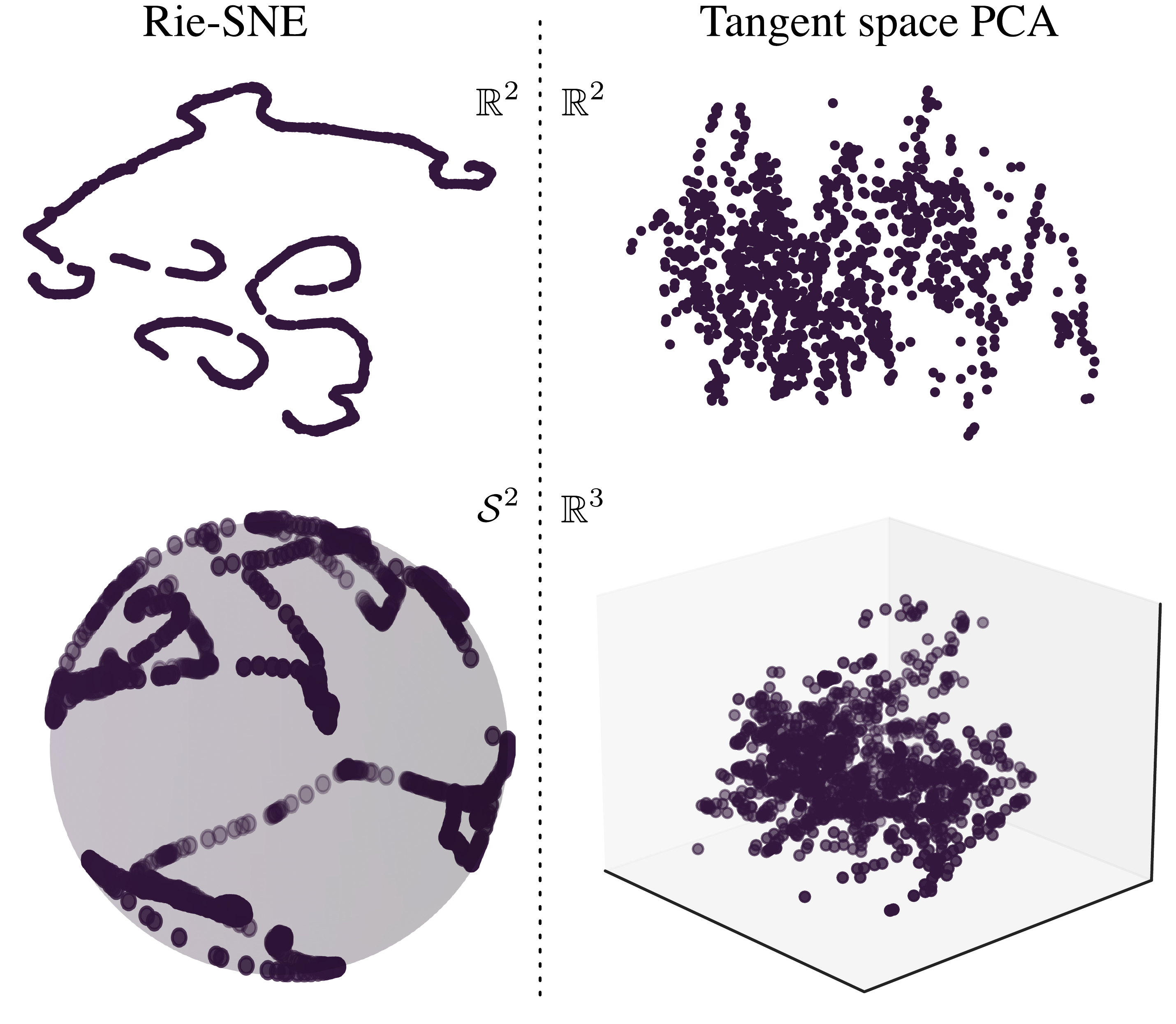}
      \caption{Embeddings of symmetric positive definite matrices using Rie-SNE (left) and tangent space PCA (right). The top row show two-dimensional Euclidean embeddings, while bottom row show spherical and $\mathbb{R}^3$ embeddings, respectively. In both cases Rie-SNE recovers a one-dimensional signal matching the underlying time series, while tangent space PCA does not.}
      \label{fig:crypto}
    \end{figure}

    Following Mallasto et al.\@ \cite{mallasto:2018} we consider the price of 10 popular crypto-currencies\footnote{Bitcoin, Dash, Digibyte, Dogecoin, Litecoin, Vertcoin, Stellar, Monero, Ripple, and Verge.} over the time period \textit{2.12.2014 --- 15.5.2018}. As is common in economy \cite{wilson2010generalised} the relationship between prices is captured by a $10 \times 10$ covariance matrix constructed from the past 20 days. This gives rise to a time series of covariance matrices, each of which reside on the cone of symmetric positive definite matrices. We provide visualizations of the data in Fig.~\ref{fig:crypto}. Rie-SNE is used to produce visualizations on both the plane $\mathbb{R}^2$ and the sphere $\mathcal{S}^2$, showing a one-dimensional structure capturing the time-evolution behind the data. In contrast, tangent space PCA produce $\mathbb{R}^2$ and $\mathbb{R}^3$ visualizations showing little to no structure in the embeddings.
    
\section{Conclusions}
In this paper, we presented a new type of visualization technique, \emph{Rie-SNE}, that is aimed at data residing on Riemannian manifolds, such as spheres. It is a SNE-based technique that can additionally produce different kinds of low-dimensional embeddings depending on user preference and the curvature of the original data manifold. We compare Rie-SNE to a standard technique when it comes to visualizing non-euclidean data, which is to perform PCA on the data mapped to tangent space. The results are promising, and we believe this technique could have some merit. For future work we would like to see this taken further: it would be interesting to see more visualizations of data mapped to well-known manifolds, other than those that were used in this paper, and it would especially be interesting to try this out on some non-standard manifolds. In the case of non-standard manifolds, computing geodesics is likely non-trivial, such that some approximation techniques might need to be developed. We hope that the present work also paves the way for other visualization tools for Riemannian data in order to support investigators relying on geometric models. 

\section*{Acknowledgements}
SH was supported by research grants (15334, 42062) from VILLUM FONDEN. This project has also received funding from the European Research Council (ERC) under the European Union's Horizon 2020 research and innovation programme (grant agreement n\textsuperscript{o} 757360).

\bibliographystyle{unsrt}
\bibliography{bibliography/refs}

\begin{thebibliography}{10}

\bibitem{gauss2005general}
Karl~Friedrich Gauss and Peter Pesic.
\newblock {\em General investigations of curved surfaces}.
\newblock Courier Corporation, 2005.

\bibitem{sne}
Geoffrey~E Hinton and Sam Roweis.
\newblock Stochastic neighbor embedding.
\newblock In S.~Becker, S.~Thrun, and K.~Obermayer, editors, {\em Advances in
  Neural Information Processing Systems}, volume~15, pages 857--864. MIT Press,
  2003.

\bibitem{vanderplas2016python}
Jake VanderPlas.
\newblock {\em Python data science handbook: Essential tools for working with
  data}.
\newblock " O'Reilly Media, Inc.", 2016.

\bibitem{asimov1985grand}
Daniel Asimov.
\newblock The grand tour: a tool for viewing multidimensional data.
\newblock {\em SIAM journal on scientific and statistical computing},
  6(1):128--143, 1985.

\bibitem{jolliffe2002principal}
Ian~T Jolliffe.
\newblock {\em Principal component analysis for special types of data}.
\newblock Springer, 2002.

\bibitem{gplvm}
Neil Lawrence.
\newblock Probabilistic non-linear principal component analysis with gaussian
  process latent variable models.
\newblock {\em The Journal of Machine Learning Research}, 6:1783--1816, 2005.

\bibitem{feldager:icann:2021}
Cilie~W. Feldager, S{\o}ren Hauberg, and Lars~Kai Hansen.
\newblock Spontaneous symmetry breaking in data visualization.
\newblock In {\em International Conference on Artificial Neural Networks
  (ICANN)}, 2021.

\bibitem{isomap}
Joshua~B Tenenbaum, Vin~de Silva, and John~C Langford.
\newblock A global geometric framework for nonlinear dimensionality reduction.
\newblock {\em science}, 290(5500):2319--2323, 2000.

\bibitem{lle}
Sam~T Roweis and Lawrence~K Saul.
\newblock Nonlinear dimensionality reduction by locally linear embedding.
\newblock {\em science}, 290(5500):2323--2326, 2000.

\bibitem{laplacian}
Mikhail Belkin and Partha Niyogi.
\newblock Laplacian eigenmaps for dimensionality reduction and data
  representation.
\newblock {\em Neural computation}, 15(6):1373--1396, 2003.

\bibitem{vanDerMaaten2008}
Laurens van~der Maaten and Geoffrey Hinton.
\newblock Visualizing data using {t-SNE}.
\newblock {\em Journal of Machine Learning Research}, 9:2579--2605, 2008.

\bibitem{mardia2000directional}
Kanti~V Mardia, Peter~E Jupp, and KV~Mardia.
\newblock {\em Directional statistics}, volume~2.
\newblock Wiley Online Library, 2000.

\bibitem{kurz2013recursive}
Gerhard Kurz, Igor Gilitschenski, and Uwe~D Hanebeck.
\newblock Recursive nonlinear filtering for angular data based on circular
  distributions.
\newblock In {\em 2013 American Control Conference}, pages 5439--5445. IEEE,
  2013.

\bibitem{hauberg:SN:2018}
S{\o}ren Hauberg.
\newblock Directional statistics with the spherical normal distribution.
\newblock In {\em Proceedings of FUSION 2018}, 2018.

\bibitem{Kendall:BLMS:1984}
D.G. Kendall.
\newblock Shape manifolds, {Procrustean} metrics, and complex projective
  spaces.
\newblock {\em Bulletin of the London Mathematical Society}, 16(2):81--121,
  1984.

\bibitem{Freifeld:ECCV:2012}
O.~Freifeld and M.J. Black.
\newblock {L}ie bodies: A manifold representation of {3D} human shape.
\newblock In {A. Fitzgibbon et al. (Eds.)}, editor, {\em European Conference on
  Computer Vision (ECCV)}, Part I, LNCS 7572, pages 1--14. Springer-Verlag,
  2012.

\bibitem{Srivastava:PAMI:2005}
Anuj Srivastava, Shantanu~H Joshi, Washington Mio, and Xiuwen Liu.
\newblock Statistical shape analysis: Clustering, learning, and testing.
\newblock {\em IEEE Transactions on Pattern Analysis and Machine Intelligence
  (TPAMI)}, 27(4):590--602, 2005.

\bibitem{Younes:ICV:2012}
Laurent Younes.
\newblock Spaces and manifolds of shapes in computer vision: An overview.
\newblock {\em Image and Vision Computing}, 30(6):389--397, 2012.

\bibitem{kurtek:pami:2012}
Sebastian Kurtek, Eric Klassen, John~C Gore, Zhaohua Ding, and Anuj Srivastava.
\newblock Elastic geodesic paths in shape space of parameterized surfaces.
\newblock {\em IEEE Transactions on Pattern Analysis and Machine Intelligence
  (TPAMI)}, 34(9):1717--1730, 2012.

\bibitem{lenglet:eccv:2004}
C.~Lenglet, R.~Deriche, and O.D. Faugeras.
\newblock Inferring white matter geometry from diffusion tensor {MRI}:
  Application to connectivity mapping.
\newblock In {\em European Conference on Computer Vision (ECCV)}, volume 3024
  of {\em Lecture Notes in Computer Science}, pages 127--140, 2004.

\bibitem{Pennec:IJCV:2006}
Xavier Pennec, Pierre Fillard, and Nicholas Ayache.
\newblock A {R}iemannian framework for tensor computing.
\newblock {\em International Journal of Computer Vision (IJCV)}, 66(1):41--66,
  2006.

\bibitem{wang2014tracking}
Yuanxiang Wang, Hesamoddin Salehian, Guang Cheng, and Baba~C Vemuri.
\newblock Tracking on the product manifold of shape and orientation for
  tractography from diffusion {MRI}.
\newblock In {\em IEEE Conference on Computer Vision and Pattern Recognition
  (CVPR)}, pages 3051--3056, 2014.

\bibitem{Tuzel:ECCV:2006}
O.~Tuzel, F.~Porikli, and P.~Meer.
\newblock Region covariance: A fast descriptor for detection and
  classification.
\newblock In {\em European Conference on Computer Vision (ECCV)}, pages
  589--600. Springer, 2006.

\bibitem{Porikli:CVPR:2006}
F.~Porikli, O.~Tuzel, and P.~Meer.
\newblock Covariance tracking using model update based on {L}ie algebra.
\newblock In {\em IEEE Conference on Computer Vision and Pattern Recognition
  (CVPR)}, volume~1, pages 728--735, 2006.

\bibitem{freifeld:cvpr:2014}
Oren Freifeld, S{\o}ren Hauberg, and Michael~J. Black.
\newblock Model transport: Towards scalable transfer learning on manifolds.
\newblock In {\em IEEE Conference on Computer Vision and Pattern Recognition
  (CVPR)}, 2014.

\bibitem{turaga}
Pavan~K. Turaga, Ashok Veeraraghavan, Anuj Srivastava, and Rama Chellappa.
\newblock Statistical computations on {G}rassmann and {S}tiefel manifolds for
  image and video-based recognition.
\newblock {\em {IEEE} Transactions on Pattern Analysis and Machine Intelligence
  (TPAMI)}, 33(11):2273--2286, 2011.

\bibitem{Cetingul:cvpr:2009}
H.E. \c{C}etingul and R.~Vidal.
\newblock Intrinsic mean shift for clustering on stiefel and grassmann
  manifolds.
\newblock In {\em IEEE Conference on Computer Vision and Pattern Recognition
  (CVPR)}, pages 1896--1902, 2009.

\bibitem{said:eusipco:2007}
Salem Said, Nicolas Courty, Nicolas Le~Bihan, and Stephen~J Sangwine.
\newblock Exact principal geodesic analysis for data on {SO(3)}.
\newblock In {\em Proceedings of the 15th European Signal Processing
  Conference}, pages 1700--1705, 2007.

\bibitem{Hauberg:IMAVIS:2011}
S{\o}ren Hauberg, Stefan Sommer, and Kim~S. Pedersen.
\newblock Natural metrics and least-committed priors for articulated tracking.
\newblock {\em Image and Vision Computing}, 30(6-7):453--461, 2012.

\bibitem{spatial_priors:hauberg_et_al10}
S{\o}ren Hauberg, Stefan Sommer, and Kim~Steenstrup Pedersen.
\newblock {Gaussian-like Spatial Priors for Articulated Tracking}.
\newblock In K.~Daniilidis, P.~Maragos, and N.~Paragios, editors, {\em ECCV},
  volume 6311 of {\em LNCS}, pages 425--437. Springer, 2010.

\bibitem{hauberg:mukf}
S{\o}ren Hauberg, Fran\c{c}ois Lauze, and Kim~Steenstrup Pedersen.
\newblock {Unscented Kalman Filtering on Riemannian Manifolds}.
\newblock {\em Journal of Mathematical Imaging and Vision}, 2011.

\bibitem{gilitschenski2015unscented}
Igor Gilitschenski, Gerhard Kurz, Simon~J Julier, and Uwe~D Hanebeck.
\newblock Unscented orientation estimation based on the bingham distribution.
\newblock {\em IEEE Transactions on Automatic Control}, 61(1):172--177, 2015.

\bibitem{jaquier2022geometry}
No{\'e}mie Jaquier, Viacheslav Borovitskiy, Andrei Smolensky, Alexander
  Terenin, Tamim Asfour, and Leonel Rozo.
\newblock Geometry-aware bayesian optimization in robotics using riemannian
  mat{\'e}rn kernels.
\newblock In {\em Conference on Robot Learning}, pages 794--805. PMLR, 2022.

\bibitem{mathieu2019continuous}
Emile Mathieu, Charline Le~Lan, Chris~J Maddison, Ryota Tomioka, and Yee~Whye
  Teh.
\newblock Continuous hierarchical representations with poincar{\'e} variational
  auto-encoders.
\newblock {\em Advances in neural information processing systems}, 32, 2019.

\bibitem{pennec}
Xavier Pennec.
\newblock {Intrinsic Statistics on Riemannian Manifolds: Basic Tools for
  Geometric Measurements}.
\newblock {\em {Journal of Mathematical Imaging and Vision}}, 25(1):127--154,
  2006.

\bibitem{fletcher2004principal}
P.~Thomas Fletcher, Conglin Lu, Stephen~M. Pizer, and Sarang Joshi.
\newblock {Principal Geodesic Analysis for the study of Nonlinear Statistics of
  Shape}.
\newblock {\em IEEE Transactions on Medical Imaging (TMI)}, 23(8):995--1005,
  2004.

\bibitem{sommer:eccv10}
Stefan Sommer, Fran\c{c}ois Lauze, S{\o}ren Hauberg, and Mads Nielsen.
\newblock {Manifold Valued Statistics, Exact Principal Geodesic Analysis and
  the Effect of Linear Approximations}.
\newblock In K.~Daniilidis, P.~Maragos, , and N.~Paragios, editors, {\em ECCV
  '10: Proceedings of the 11th European Conference on Computer Vision}, Lecture
  Notes in Computer Science, pages 43--56. Springer, Heidelberg, September
  2010.

\bibitem{huckemann}
S.~Huckemann, T.~Hotz, and A.~Munk.
\newblock Intrinsic shape analysis: geodesic {PCA} for {R}iemannian manifolds
  modulo isometric {L}ie group actions.
\newblock {\em Statistica Sinica}, 20(1):1--58, 2010.

\bibitem{panaretos:jasa:2014}
Victor~M. Panaretos, Tung Pham, and Zhigang Yao.
\newblock Principal flows.
\newblock {\em Journal of the American Statistical Association (JASA)},
  109(505):424--436, 2014.

\bibitem{jung:biometrika:2012}
Sungkyu Jung, Ian~L Dryden, and JS~Marron.
\newblock Analysis of principal nested spheres.
\newblock {\em Biometrika}, 99(3):551--568, 2012.

\bibitem{mallasto:2018}
Anton Mallasto, S{\o}ren Hauberg, and Aasa Feragen.
\newblock Probabilistic riemannian submanifold learning with wrapped gaussian
  process latent variable models.
\newblock In {\em Proceedings of the 19th international Conference on
  Artificial Intelligence and Statistics (AISTATS)}, 2018.

\bibitem{hauberg:tpami:princurve}
S{\o}ren Hauberg.
\newblock Principal curves on riemannian manifolds.
\newblock {\em IEEE Transactions on Pattern Analysis and Machine Intelligence
  (TPAMI)}, 2015.

\bibitem{doCarmoRiemannian}
M.P. do~Carmo.
\newblock {\em Riemannian Geometry}.
\newblock Mathematics (Boston, Mass.). Birkh{\"a}user, 1992.

\bibitem{journals/bstj/Shannon48}
Claude~E. Shannon.
\newblock A mathematical theory of communication.
\newblock {\em Bell Syst. Tech. J.}, 27(3):379--423, 1948.

\bibitem{tsneOptimization}
Laurens van~der Maaten.
\newblock Accelerating t-sne using tree-based algorithms.
\newblock {\em J. Mach. Learn. Res.}, 15(1):3221–3245, January 2014.

\bibitem{hsu2002stochastic}
E.P. Hsu and American~Mathematical Society.
\newblock {\em Stochastic Analysis on Manifolds}.
\newblock Graduate studies in mathematics. American Mathematical Society, 2002.

\bibitem{kalatzis2020variational}
Dimitris Kalatzis, David Eklund, Georgios Arvanitidis, and Søren Hauberg.
\newblock Variational autoencoders with riemannian brownian motion priors,
  2020.

\bibitem{vptrees}
Peter~N. Yianilos.
\newblock Data structures and algorithms for nearest neighbor search in general
  metric spaces.
\newblock In {\em SODA}, pages 311--321, 1993.

\bibitem{barnes86a}
J.~E. Barnes and P.~Hut.
\newblock A hierarchical {O}(n-log-n) force calculation algorithm.
\newblock {\em Nature}, 324:446, 1986.

\bibitem{lecun-mnisthandwrittendigit-2010}
Yann LeCun and Corinna Cortes.
\newblock {MNIST} handwritten digit database.
\newblock {\em Online}, 2010.

\bibitem{wilson2010generalised}
Andrew~Gordon Wilson and Zoubin Ghahramani.
\newblock Generalised wishart processes.
\newblock {\em arXiv preprint arXiv:1101.0240}, 2010.

\end{thebibliography}


\begin{algorithm} 
    \SetAlgoLined
    \caption{Computing conditional probabilities $p_{j|i}$}\label{alg:main}
    \small
    \KwData{Geodesic distance matrix: $D$,\\
    \quad \quad \hspace{0.15em} \ Riemannian volume measure ratio matrix: $\mathcal{H}_0$,\\
    \quad \quad \hspace{0.15em} \ Data space dimensionality: $dim$,\\
    \quad \quad \hspace{0.15em} \ Desired perplexity: $desired\_perplexity$\\
    }
    \KwResult{$n \times n$ matrix of conditional probabilities: $P$}
    \Begin{
    $P \leftarrow{} zeros(n, n)$\\
    $binary\_search\_steps \leftarrow{} 100$\\
    $perplexity\_tolerance \leftarrow{} 10^{-5}$\\
    \For{i=0 \textbf{\text{to}} n}{
        $t\_min \leftarrow{} -\infty$\\
        $t\_max \leftarrow{} \infty$\\
        $t \leftarrow{} 1.0$\\
        \For{l=0 \textbf{\text{to}} $binary\_search\_steps$}{
            $row\_sum \leftarrow{} 0.0$\\
           \For{j=0 \textbf{\text{to}} $n$}{
                $P[i, j] \leftarrow{} \left( 2 \pi t \right)^{-\frac{dim}{2}} \cdot \mathcal{H}_{0}[i,j] \cdot \exp \left( - \frac{D[i, j]^{2}}{2t}\right)$\\
                $row\_sum \leftarrow{} row\_sum + P[i, j]$
            }
            $entropy \leftarrow{} 0.0$\\
           \For{j=0 \textbf{\text{to}} $n$}{
                $P[i, j] \leftarrow{} \frac{P[i,j]}{row\_sum} $\\
                \If{$P[i,j] \neq 0.0$}{
                    $entropy \leftarrow{} entropy + P[i,j] \cdot \log_2(P[i,j])$
                }
            }
            
            $entropy\_diff \leftarrow{} - entropy - \log_2(desired\_perplexity)$\\
            
            \If{$|entropy\_diff| \leq perplexity\_tolerance$}{
                $break$
            }
            
            \eIf{$entropy\_diff < 0.0$}{
                $t\_min \leftarrow{} t$\\
                \eIf{$t\_max == \infty$}{
                    $t \leftarrow{} 2 \cdot t$
                }
                {
                    $t \leftarrow{} \frac{t + t\_max}{2.0}$
                }
            }
            {
                $t\_max \leftarrow{} t$\\
                \eIf{$t\_min == -\infty$}{
                    $t \leftarrow{} \frac{t}{2}$
                }
                {
                    $t \leftarrow{} \frac{t + t\_min}{2.0}$
                }
            }
            
        }
    }
    \Return $P$
    }
\end{algorithm}

\end{document}